\documentclass[10pt,twocolumn,letterpaper]{article}

\usepackage{cvpr}              
\definecolor{cvprblue}{rgb}{0.21,0.49,0.74}
\usepackage[pagebackref,breaklinks,colorlinks,allcolors=cvprblue]{hyperref}
\usepackage{multirow}
\usepackage{makecell}
\usepackage{tabularray}
\usepackage{colortbl}
\usepackage{graphicx}
\usepackage{bbding}
\usepackage{pifont}

\usepackage[accsupp]{axessibility}

\def\paperID{19025} 
\def\confName{CVPR}
\def\confYear{2026}

\vspace{-1.0cm}
\title{Pano360: Perspective to Panoramic Vision with Geometric Consistency}
\author{
Zhengdong Zhu$^{1*}$ \quad Weiyi Xue$^{2*}$ \quad Zuyuan Yang$^{3}$ \vspace{0.2cm} \\
Wenlve Zhou$^{1}$ \quad Zhiheng Zhou$^{1\dagger}$ \\[0.2cm] %
$^1$South China University of Technology \quad $^2$Tongji University \quad \vspace{0.2cm} \\ 
$^3$Guangdong University of Technology
}

\newcommand\blfootnote[1]{%
  \begingroup
  \renewcommand\thefootnote{}\footnote{#1}%
  \addtocounter{footnote}{-1}%
  \endgroup
}

\begin{document}
\maketitle
\blfootnote{$^*$ Equal Contribution. $^\dagger$ Corresponding Author.}

\begin{abstract}
Prior panorama stitching approaches heavily rely on pairwise feature correspondences and are unable to leverage geometric consistency across multiple views. This leads to severe distortion and misalignment, especially in challenging scenes with weak textures, large parallax, and repetitive patterns.
Given that multi-view geometric correspondences can be directly constructed in 3D space, making them more accurate and globally consistent, we extend the 2D alignment task to the 3D photogrammetric space. 
We adopt a novel transformer-based architecture to achieve 3D awareness and aggregate global information across all views. It directly utilizes camera poses to guide image warping for global alignment in 3D space and employs a multi-feature joint optimization strategy to compute the seams.
Additionally, to establish an evaluation benchmark and train our network, we constructed a large-scale dataset of real-world scenes. Extensive experiments show that our method significantly outperforms existing alternatives in alignment accuracy and perceptual quality. Resources are available at \url{https://github.com/KiMomota/Pano360}
\end{abstract}

\section{Introduction}
\label{sec:intro}

With the increasing demand for immersive perceptual experiences and holistic scene understanding, the panorama provides a large field-of-view (FoV), which has drawn growing attention from the research community.
Panorama stitching techniques are indispensable for emerging applications such as autonomous driving~\cite{kong2025multi} and virtual reality~\cite{linhqgs}. They also facilitate downstream tasks including 3D Gaussian splatting~\cite{chen2025splatter} and text-to-panorama generation~\cite{zhang2024taming}.

However, creating a globally consistent and visually pleasing panorama, especially from a large number of input images, remains a significant challenge. Traditionally, panorama stitching methods primarily rely on matching handcrafted features to estimate pairwise homography, such as keypoints, line segments, curves, and facets~\cite{wang2024efficient, jia2021leveraging, du2022geometric, li2019local}.
However, this pairwise approach is prone to accumulating projection errors, particularly when aligning multiple images, resulting in severe distortions in the remaining images.
Furthermore, when dealing with large-scale real-world scenarios, this issue is exacerbated by weak textures, large parallax, and repetitive patterns. The scarcity of reliable feature matches often leads to homography estimation failure. As traditional geometric feature-based methods suffer from error accumulation and perform poorly in these challenging scenarios, it becomes essential to incorporate stronger geometric priors for panorama stitching.

\begin{figure}[t]
    \centering
    \includegraphics[width=1.0\linewidth]{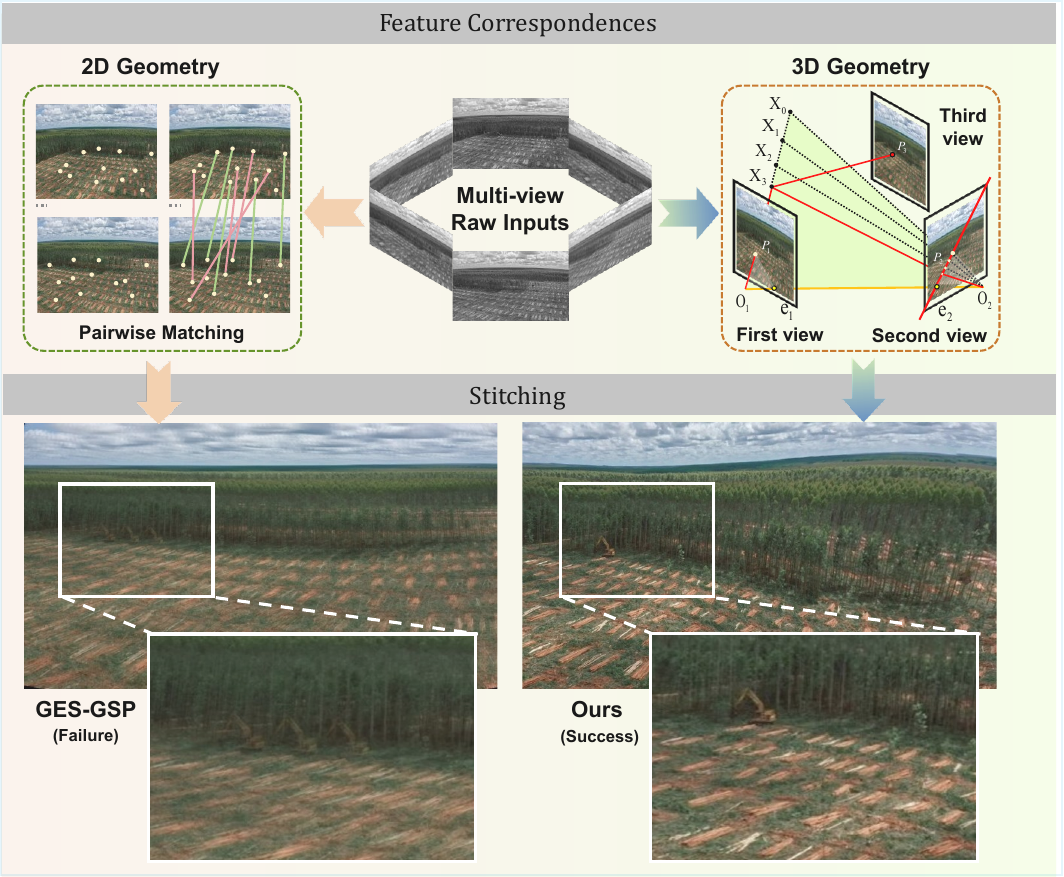} 
    \caption{\textbf{Comparison in repetitive patterns.} GES-GSP~\cite{du2022geometric} suffers from inaccurate pairwise feature correspondences, causing severe artifacts. Leveraging multi-view geometric consistency in 3D space effectively filters unreliable matches, enabling robust alignment. Our method achieves a 97.8\% success rate in challenging scenes with weak textures, large parallax, and repetitive patterns.}
    \label{fig:teaser}
\end{figure}

Consequently, recent works such as UDIS~\cite{nie2021unsupervised} and UDIS2~\cite{nie2023parallax} have employed Convolutional Neural Networks (CNNs) for image stitching using end-to-end learning techniques.
Although they demonstrate superior performance in specific scenes where geometric features are scarce, these networks are fundamentally limited to pairwise image stitching and require complex post-processing for multi-image alignment. These limitations severely hinder their real-world applicability.

Almost all existing panorama stitching methods, including traditional feature-based and learning-based methods are confined to establishing pairwise correspondences in 2D space. They aim to achieve the visually seamless panorama but ignore the underlying 3D projective geometry required for global coherence. This leads to severe distortion and misalignment due to inaccurate and inconsistent feature correspondences, as shown in~\cref{fig:teaser}. To overcome these aforementioned limitations, we introduce Pano360, a novel network that departs from the conventional pairwise pipeline. It directly aligns images in 3D photogrammetric space, explicitly preserving the globally geometric consistency.

Specifically, inspired by the large vision models~\cite{kirillov2023segment,yang2024depth,simeoni2025dinov3}, such as VGGT~\cite{wang2025vggt}, which is trained with 3D supervision and thus exhibits an inherent 3D awareness of feature correspondences. Therefore, we adapt its architecture to aggregate
features across all input images and achieve 3D awareness of the scene. 
Our network directly utilizes camera poses to guide image warping in 3D space and selects the optimal projection function from the aggregated multi-image features. Pano360 supports inputs ranging from a few to hundreds of images and generates a geometric consistent panorama. In addition, to ensure seamless transitions, we propose a multi-feature joint optimization strategy for seam identification in complex overlaps. By integrating color, gradient, and texture terms from all images, our method eliminates artifacts while significantly reducing costly pairwise computations, achieving up to a 10 times speedup over prior pairwise methods in large-scale scenes.

To effectively train our transformer network and establish a robust evaluation benchmark, a large-scale panoramic dataset of high-quality, real-world images is required. However, existing datasets~\cite{nie2021unsupervised,nie2025stabstitch++} are typically synthetic, lack scene diversity, and primarily consist of image pairs without viewpoint variation. Therefore, we construct a new large-scale panorama dataset comprising over 14,400 images from 200 diverse real-world scenes with challenging conditions such as weak texture, varied illumination, and extreme weather. Each scene covers a full 360° FoV and all images are annotated with ground-truth camera parameters.

To summarize, we make the following contributions:
\begin{itemize}
    \item We introduce Pano360, a novel pipeline that preserves the 3D geometric consistency for panorama stitching. Given inputs ranging from a few to hundreds of images, it directly utilizes camera poses to guide image alignment in 3D space and seamlessly blends them into a panorama.
    
    \item We construct a large-scale, real-world dataset comprising 200 diverse scenes, each image from a wide range of viewpoints that collectively cover a 360° FoV.
    
    \item Comprehensive experiments demonstrate that Pano360 achieves state-of-the-art performance in a variety of challenging environments, significantly outperforming previous approaches in terms of both alignment accuracy and perceptual quality.
\end{itemize}

\section{Related Work}
\label{sec:related work}

Computer vision techniques have made significant progress in panorama stitching. Most of the contributions in the past concentrated on pairwise perspective image stitching.

\noindent\textbf{Feature Association} is a classic computer vision problem involving feature extraction and matching from a set of images or video frames. It has been widely applied in geometric vision tasks, such as SLAM and SfM~\cite{wang2024vggsfm}. To establish geometric relationships between different views, feature descriptors like SIFT~\cite{lowe2004distinctive}, SURF~\cite{bay2006surf}, ORB~\cite{rublee2011orb}, the transformer-based LoFTR~\cite{wang2024efficient} are commonly used. Outlier filtering, typically using RANSAC, is essential for robust homography estimation. When keypoints are scarce, other methods leverage geometric primitives such as line segments, curves, and facets for reliable matching~\cite{jia2021leveraging, du2022geometric, li2019local}. Alternatively, optical flow~\cite{xu2022gmflow} can estimate correspondences by extracting motion-consistent trajectories. However, panorama stitching suffers performance drop from feature detection  and matching, when the scenes contain weak textures, large parallax, or repetitive patterns.

\noindent\textbf{Geometric Alignment} aims to recover the consistent geometric relationship among multiple overlapping images of a scene. Generally, a homography matrix is used to describe the geometric relationship between perspective images. However, a planar homography transformation is inadequate for scenes with depth variation and parallax. Thus, multi-homography~\cite{lee2020warping,song2021end}, mesh-based~\cite{zhang2025adapting, nie2021depth,liao2019single}, and global-aware~\cite{jiang2024multispectral,li2024automatic} transformations are often used for image alignment. In end-to-end image stitching, deep learning-based approaches have recently made significant progress. For instance, UDIS~\cite{nie2021unsupervised} and UDIS2~\cite{nie2023parallax} directly estimate aligned images using an iterative mesh refinement strategy in a VGG network. For stable stitching results with reduced jitter and distortions, StabStitch~\cite{nie2024eliminating} and StabStitch++~\cite{nie2025stabstitch++} optimize deformation trajectories within the spatio-temporal domain. Although these CNN-based methods outperform traditional approaches in specific scenes, certain constraints significantly impair their generalization capabilities, severely limiting their practical applicability.

\begin{figure*}[t]
    \centering

    \includegraphics[width=0.97\textwidth]{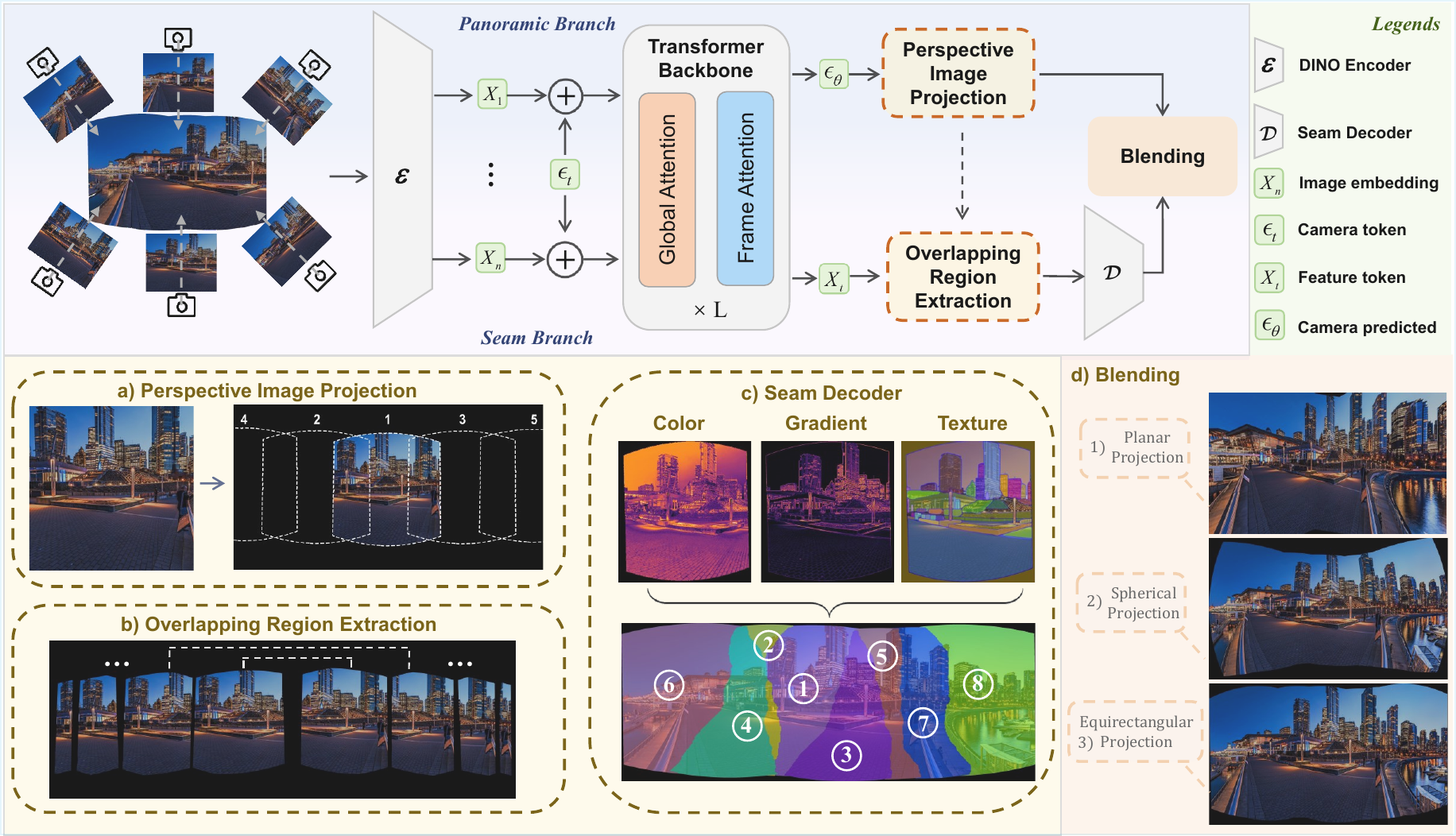} 
    
    \caption{\textbf{Architecture Overview.} (a) Perspective images are projected onto a common panoramic coordinate system using camera parameters. (b) Overlapping regions are extracted from the globally aligned images (a single image may overlap multiple neighbors). (c) The seam decoder is supervised with three weight maps and generates the seam mask for each image. (d) The final panorama is blended using the seam mask and the globally aligned images, supporting various projection formats.}
    
    \label{fig:architecture} 
\end{figure*}

\noindent\textbf{Seam Detection.} A crucial step in panorama stitching is identifying an optimal seam within overlapping regions to hide transition boundaries~\cite{brown2007automatic}. The goal is to ensure smooth transitions between images, thereby eliminating artifacts and enhancing visual consistency. Typically, this is formulated as an energy minimization problem. The energy function often considers multiple terms, such as differences in color, gradient~\cite{nie2023parallax}, edges, or even quaternion-domain color features~\cite{li2024automatic,jiang2024multispectral}. The final seam is then determined by minimizing this energy using algorithms like graph-cut or dynamic programming~\cite{boykov2002fast}. However, existing methods are often computationally expensive and limited to pairwise image processing, which is prone to converging to locally optimal solutions in complex, multi-image overlaps.

\section{Method}
We introduce Pano360, a transformer network that ingests a batch of input images and predicts all required quantities for panorama stitching. We begin by presenting the problem definition in~\cref{sec:Problem definition}, followed by an overview of the overall architecture in~\cref{sec:Feature backbone}. We then detail the projection head in~\cref{sec:Projection Head} and the seam
head in~\cref{sec:Seam head}, concluding with the training process in~\cref{sec:Training}.

\begin{figure*}[t] 
    \centering
    
    \includegraphics[width=1.0\textwidth]{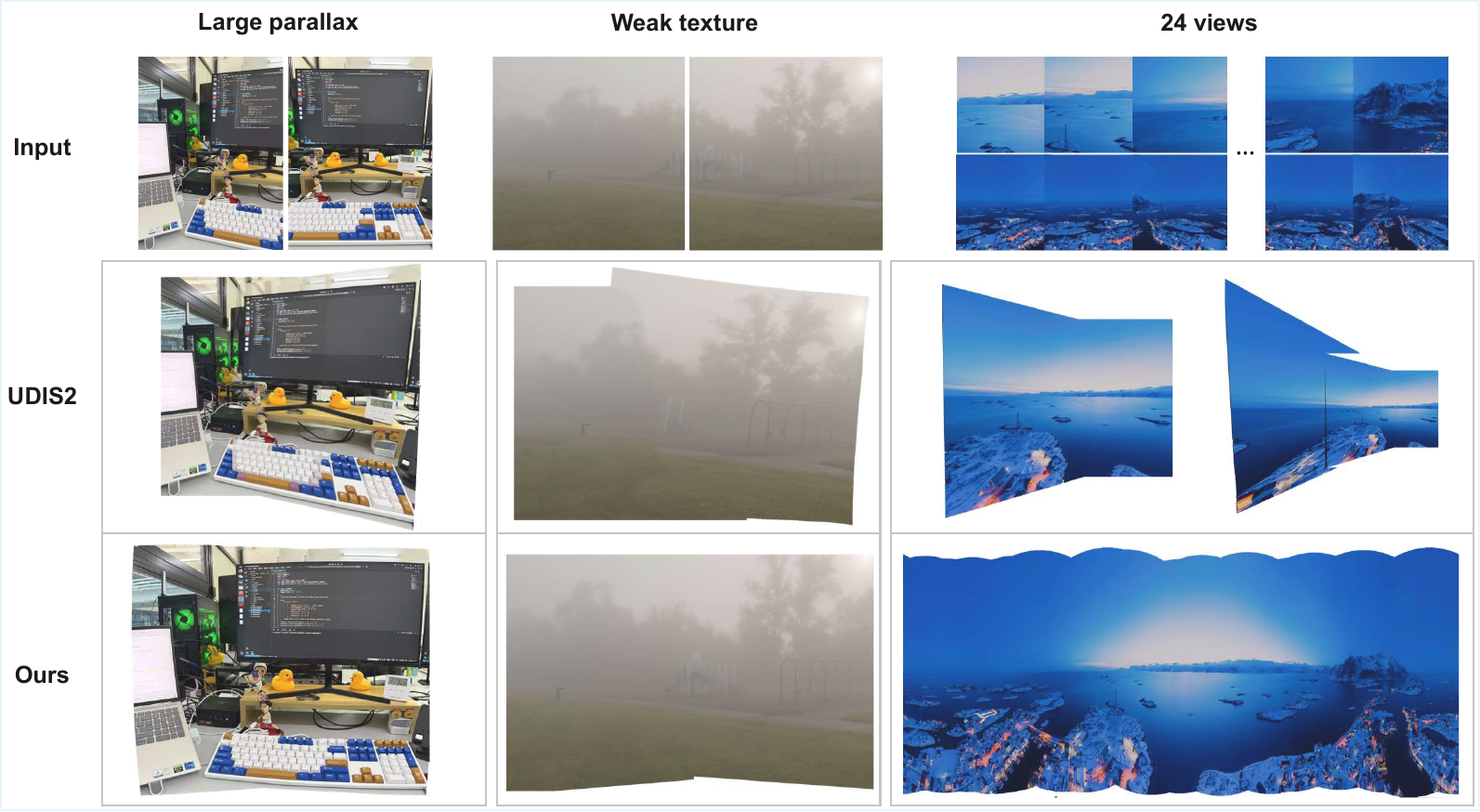}
    
    \caption{\textbf{Qualitative comparison of our method Pano360 to UDIS2 on challenging real-world scenarios.} As shown in the first column, our method successfully handles the large parallax scene, while UDIS2 suffers from ghosting and misalignment. In the second column, our method stitches the images in the weak-textured fog scene, while UDIS2 fails. The third column presents a challenging case with repetitive patterns, unusual lighting, and a large FoV. Our method successfully recovers a panorama from 24 frames, whereas UDIS2 is limited to pairwise processing and suffers from accumulated errors, leading to severe geometric distortion and failure.}
    
    \label{fig:comparison3}
\end{figure*}

\subsection{Problem definition}
\label{sec:Problem definition}

Given a set of $N$ partially overlapping images $\{I_i\}_{i=1}^N$, where each image $I_i \in \mathbb{R}^{H \times W \times 3}$, the goal of panorama stitching is to generate a single, seamless panoramic image $\mathcal{I}$. This process requires estimating a set of parameters for each input image that defines its transformation and contribution to the final panorama.

Specifically, for each image $I_i$, the goal is to find a warping function $\mathcal{W}_i$ that maps pixels to the panoramic coordinates. This warp can be decomposed into two components: a globally projective transformation $P_i$ and a local deformation field $W_i$. The global projection $P_i$ performs a global alignment of the image, while the local mesh warp $W_i$ applies fine-grained, pixel-level corrections to handle parallax distortions. The complete warp for a pixel at coordinate $\mathbf{u}$ in image $I_i$ is defined as:
\begin{equation}
\mathcal{W}_i(\mathbf{u}) = P_i(\mathbf{u}) + W_i(\mathbf{u}).
\label{eq:warp}
\end{equation}

To create a seamless composite, the seam mask $M_i$  selects which pixels from each warped image are used in the final panorama. Therefore, the network $f$ ingests the set of images $\{I_i\}_{i=1}^N$ as input and jointly predicts all the required quantities:
\begin{equation}
    f(\{I_i\}_{i=1}^N) = \{ P_i, W_i, M_i \}_{i=1}^N.
    \label{eq:problem_definition}
\end{equation}

Additionally, to correct the color and lighting variations between images, image blending methods such as feathering, Poisson, and multiband blending~\cite{lin2025one} are used to generate a visually harmonious image $\mathcal{I}$. 



\subsection{Feature backbone}
\label{sec:Feature backbone}

Our network, Pano360, employs a dual-branch architecture built upon a shared feature backbone, as illustrated in~\cref{fig:architecture}. Each image $\{I_i\}_{i=1}^N$ is initially patchified and processed by a pre-trained DINO~\cite{simeoni2025dinov3} encoder. To learn globally geometric relationships across all images, a learnable camera token is prepended to the concatenated sequence of all image embeddings. 
This combined sequence is then processed by $L$ alternating-attention transformer layers from pre-trained VGGT~\cite{wang2025vggt}, which includes global attention and frame attention. 
Finally, the backbone produces two distinct outputs for the downstream branches: predicted camera tokens that contain 3D geometric correspondences for image alignment, and feature tokens that retain details for seam mask identification.

\begin{figure*}[t]
    \centering

    \includegraphics[width=1.0\linewidth]{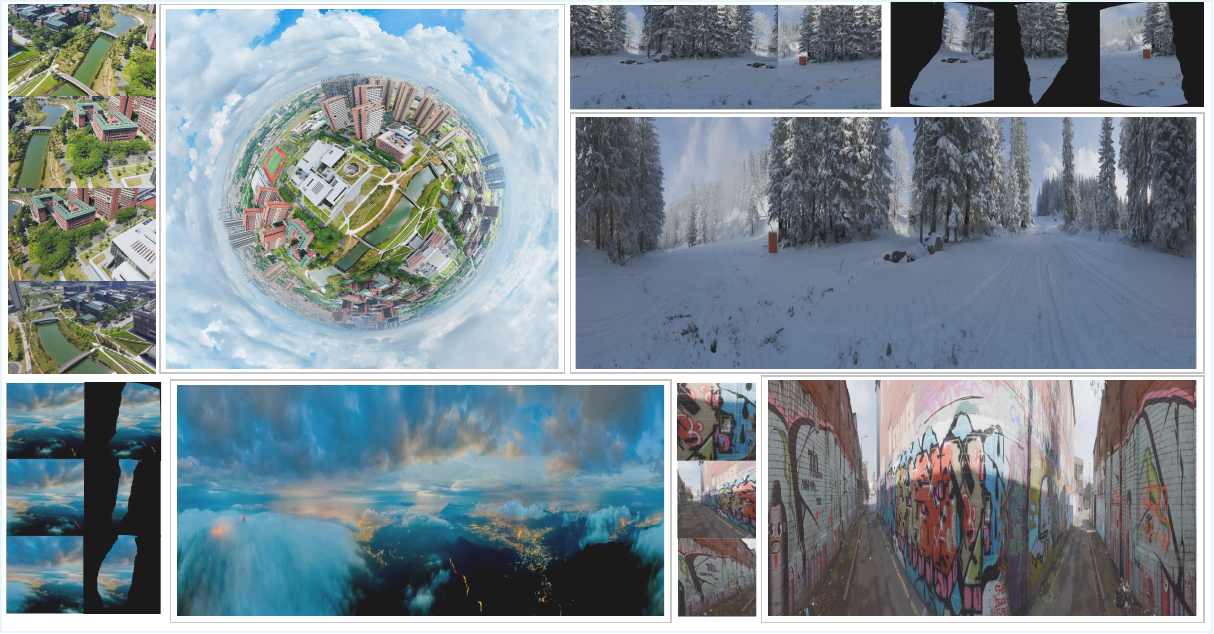} 
    
    \caption{\textbf{Visualization of the proposed Pano360 dataset.} It showcases distinct and challenging scenes. Each case presents several source images, their seam masks, and the final panorama we generated.}
    
    \label{fig:dataset_intro}
\end{figure*}

\subsection{Projection Head}
\label{sec:Projection Head}

The projection head decodes the predicted camera token into camera parameters: intrinsics $\{\mathbf{K_i}\}_{i=1}^N$ and extrinsics $\{\mathbf{R}_i, \mathbf{t}_i\}_{i=1}^N$, which collectively define the geometric relationship between all input images. Specifically, we assume that all cameras share the same focal length and the principal points are at the image center. The pose of the first camera is fixed as the reference frame of $\mathbf{R}_1 = \mathbf{I}$ and $\mathbf{t}_1 = \mathbf{0}$. 
This set of assumptions is commonly used in SfM frameworks~\cite{wang2024vggsfm}. 
To adapt to scenes captured at different resolutions, we predict the camera parameters at the scale fixed by the network. During the alignment, the intrinsics are then linearly rescaled to the original resolution.

Crucially, these camera parameters define a projection function $\mathbf{P}_i$ for each image, which maps its pixels onto a common coordinate system. Given the camera parameters, our method supports choosing arbitrary projection formats for panoramic stitching. Furthermore, the network can also adaptively predict a suitable panoramic representation format for a given set of images. As shown in \cref{fig:architecture}, our model leverages camera parameters for image alignment, which support a variety of projection formats, including planar, equirectangular, and spherical projection.

Finally, we warp each image $I_i$ using its corresponding projection function $\mathbf{P_i}$ onto the panoramic canvas to achieve a globally consistent alignment. However, for specific scenarios with large parallax and significant depth variations, we need to further compute a local mesh warp $W_i$ for each image to correct the local residual misalignments. Additional implementation details on the projection formats and the mesh warp are provided in the \textit{Supplementary Material}.

\subsection{Seam Head}
\label{sec:Seam head}

Here, we introduce how to obtain high-quality seam mask labels to train our seam head. This ensures that our network can predict the globally optimal seam for each image in a single forward pass, without requiring complex pairwise computations or post-processing.

\begin{table*}[t]
\centering
\resizebox{\textwidth}{!}{
\renewcommand{\arraystretch}{1.2}
\setlength{\tabcolsep}{6pt} 
\begin{tabular}{ccccccccccccc}
\hline
\multirow[c]{2}{*}{ Method } &
\multicolumn{4}{c}{Scene (a)} &
\multicolumn{4}{c}{Scene (b)} &
\multicolumn{4}{c}{Scene (c)} \\
\cmidrule(lr){2-5} \cmidrule(lr){6-9} \cmidrule(lr){10-13}
& $\text{QA}_{\text{q}}$$\uparrow$ & $\text{QA}_{\text{a}}$$\uparrow$ & BRIS$\downarrow$ & NIQE$\downarrow$
& $\text{QA}_{\text{q}}$$\uparrow$ & $\text{QA}_{\text{a}}$$\uparrow$ & BRIS$\downarrow$ & NIQE$\downarrow$
& $\text{QA}_{\text{q}}$$\uparrow$ & $\text{QA}_{\text{a}}$$\uparrow$ & BRIS$\downarrow$ & NIQE$\downarrow$ \\
\midrule

AutoStitch~\cite{brown2007automatic} & 3.82 & 3.20 & 40.98 & 4.55 & 3.24 & 2.74 & 43.50 & 5.94 & 3.28 & 2.81 & 49.84 & 5.01 \\

APAP~\cite{zaragoza2013projective} & 3.62 & 3.05 & 54.49 & 3.72 & 3.22 & 3.03 & 52.62 & 3.78 & 3.53 & 3.66 & 45.66 & 3.77 \\

ELA~\cite{li2017parallax} & \underline{3.88} & 3.17 & 46.15 & 3.71 & 3.35 & 2.60 & 40.69 & 5.02 & \underline{3.75} & 3.20 & \underline{41.54} & 3.88 \\

GSP~\cite{chen2016natural} & 3.87 & \underline{3.25} & 37.55 & 3.53 & 3.20 & \underline{3.45} & 40.66 & 3.96 & 3.65 & 3.57 & 42.19 & \underline{3.71} \\

GES-GSP~\cite{du2022geometric} & 3.95 & 3.20 & \underline{36.45} & \underline{3.36} & \underline{3.45} & 3.35 & \underline{32.25} & \underline{3.95} & 3.74 & \underline{3.72} & 44.22 & 3.95 \\

LPC~\cite{jia2021leveraging}$^{\ddag}$ & 3.01 & 2.88 & 57.55 & 5.86 & 2.57 & 2.68 & 55.37 & 6.81 & 2.98 & 2.21 & 59.88 & 4.96 \\

UDIS2~\cite{nie2023parallax}$^{\ddag}$ & 3.02 & 2.97 & 60.55 & 5.23 & 2.84 & 2.64 & 51.63 & 6.53 & 2.87 & 2.34 & 58.62 & 4.91 \\

\midrule
\textbf{Pano360 (Ours)} & \textbf{4.08} & \textbf{3.40} & \textbf{34.65} & \textbf{3.26} & \textbf{3.57} & \textbf{3.52} & \textbf{28.79} & \textbf{3.71} & \textbf{4.09} & \textbf{3.94} & \textbf{37.96} & \textbf{3.37} \\

\bottomrule
\end{tabular}}
\caption{\textbf{Quantitative comparison on the Pano360 Dataset.} We compare our method to different types of previous approaches and mark the top results as \textbf{best} and \underline{second best}. None of these methods were trained on the Pano360 testing dataset. Methods marked with $^{\ddag}$ only accept pairwise image inputs. All experiments were conducted on a single NVIDIA A100 GPU.}
\label{tab:Pano-360_comparison}
\end{table*}

Given a set of aligned images $\{I_i\}_{i=1}^N$ with overlapping regions, our goal is to assign a label $M(p) \in \{1,2,\dots,N\}$ for each pixel $p$ in these regions. This label indicates which images the pixel originates from. These seams are defined as the 
boundaries between regions with different labels. This problem is proposed as an energy minimization problem. The energy function is composed of two terms:
\begin{equation}
E(\mathcal{I}) = E_{l}(\mathcal{I}) + E_{c}(\mathcal{I})
\label{eq:energy}
\end{equation}

Where $E_{l}$ measures the cost of assigning a particular label to a pixel. We introduce a hard constraint to ensure pixels are only sourced from images where they are valid:

\begin{equation}
E_{l}(\mathcal{I}) = \sum_{i=1}^{N} \sum_{p \in \mathcal{I}} D_{I_i}(p)
\end{equation}
\begin{equation}
D_{I_i}(p) = \begin{cases}
0, & \text{} p \in I_i \\
\infty, & \text{} p \notin I_i \\
\end{cases}
\end{equation}

By applying a penalty to invalid assignments, the search for optimal seamline is effectively constrained to the overlapping regions.
Then the $E_{c}$ penalizes assigning different labels to adjacent pixels, encouraging the seam to be continuous and inconspicuous. It is defined as the sum of costs over all adjacent pixel pairs:

\begin{equation}
E_{c}(\mathcal{I}) = \sum_{(p,q) \in \mathcal{E}}V(M(p),M(q))
\end{equation}

where $\mathcal{E}$ is the set of all adjacent pixel pairs. The penalty function $V$ is zero if the adjacent pixels $p$ and $q$ are assigned the same label. If the labels differ, a seam is created, and the penalty is based on a pixel-wise cost function $V(\cdot)$:

\begin{equation}
V(M(p),M(q)) = \begin{cases}
C(p) + C(q), & \text{} M(p) \neq M(q) \\
0, & \text{} M(p) = M(q)
\end{cases}
\end{equation}

The pixel-wise cost $C(p)$ is designed to be high in visually complex areas and low in simple, uniform regions. Following the approach defined in~\cite{li2016optimal}, we utilize a color difference map $F_{color}(p)$, a gradient magnitude map $F_{gradient}(p)$, and a texture complexity map $F_{ratio}(p)$ to guide the seamline generation. Thus, we define this cost as a combination of color, gradient, and texture information.

\begin{equation}
C(p)=F_{color}(p) + F_{gradient}(p) \times F_{ratio}(p)
\label{eq:cost function}
\end{equation}

The cost function is guided by three maps. For any pair of overlapping images, the color difference map $F_{color}(p) = || I_i(p)-I_j(p) ||$ and the gradient magnitude map $F_{gradient}(p) = |\nabla I_i(p)| + |\nabla {I}_j(p)|$ form a baseline cost for fundamental obstacle avoidance, steering the seam away from color discontinuities and sharp object edges. Then, the gradient is modulated by the texture complexity map, which provides high-level guidance. It heavily penalizes seams in visually complex regions (especially those with parallax or depth variations) and directs them towards homogeneous areas where artifacts are less perceptible.

In this method, our method can simultaneously consider the color difference and gradient magnitude of all images within an overlapping region, while also utilizing a texture term to guide the seamline generation. This overcomes the limitation of traditional methods, which are limited to pairwise estimations and fall into local optima. We then use the computed seam mask $\hat{M_i}$ of each image, as pseudo-labels to supervise the training of the seam decoder~\cite{ranftl2021vision}.

\begin{table*}[t]
  \centering
  
\begin{minipage}[t]{0.48\textwidth}
    \centering
    \footnotesize
    \setlength{\tabcolsep}{6pt} 
    \renewcommand{\arraystretch}{1.2} 
    \begin{tabular}{ccccc}
      \toprule
      \multicolumn{1}{c}{\multirow{2}{*}{{Geometry}}} & 
      \multicolumn{1}{c}{\multirow{2}{*}{{Method}}} & 
      \multicolumn{1}{c}{\multirow{2}{*}{{Success (\%)}}} & 
      \multicolumn{1}{c}{\multirow{2}{*}{{Runtime}}} & \\
      &&&& \\[0.0 em]
      \midrule
      \checkmark & LoFTR+RANSAC~\cite{wang2024efficient} & 63.4 & $\sim$ 13s \\
      \checkmark & LightGlue+RANSAC~\cite{lindenberger2023lightglue} & 66.7 & $\sim$ \textbf{11s} \\
      \checkmark & AutoStitch~\cite{brown2007automatic} & 46.7 & $\sim$ 60s \\
      \checkmark & APAP~\cite{zaragoza2013projective} & 30.0 & $ > 300$s \\
      \checkmark & ELA~\cite{li2017parallax} & 80.1 & $\sim$ 90s \\
      \checkmark & GSP~\cite{chen2016natural} & 77.6 & $\sim$ 30s \\
      \checkmark & GES-GSP~\cite{du2022geometric} & \textbf{83.3} & $\mathbf{\sim}$ 20s \\
      \midrule
      \ding{55} & {\textbf{Ours}} & \textbf{97.8} & {$\mathbf{\sim}$ \textbf{5s}} \\
      \bottomrule
    \end{tabular}
    \caption{\textbf{Comparison of success rate and runtime on the Pano360 Dataset.} Methods marked with \checkmark denotes explicitly geometric feature-based methods (e.g., points, lines).}
    \label{tab:geometric_feature_comparison}
  \end{minipage}
  \hfill 
\begin{minipage}[t]{0.48\textwidth}
    \centering
    \footnotesize
    \setlength{\tabcolsep}{6pt} 
    \renewcommand{\arraystretch}{1.2} 
    \begin{tabular}{ccccc}
      \toprule
      \multicolumn{1}{c}{\multirow{2}{*}{{Method}}} & 
      \multicolumn{1}{c}{\multirow{2}{*}{{PSNR $\uparrow$}}} & 
      \multicolumn{1}{c}{\multirow{2}{*}{{SSIM $\uparrow$}}} & 
      \multicolumn{1}{c}{\multirow{2}{*}{{PIQE $\downarrow$}}} &
      \multicolumn{1}{c}{\multirow{2}{*}{{NIQE $\downarrow$}}} \\
      &&&& \\[0.0 em]
      \midrule
      SIFT+RANSAC~\cite{lowe2004distinctive} & 23.27 & 0.779 & 56.26 & 15.38 \\
      APAP~\cite{zaragoza2013projective} & 23.79 & 0.794 & 53.36 & 14.16 \\
      SPW~\cite{liao2019single} & 21.60 & 0.687 & 51.46 & 14.31 \\
      LPC~\cite{jia2021leveraging}$^{\ddag}$ & 22.59 & 0.736 & 52.41 & 13.43 \\
      UDIS~\cite{nie2021unsupervised}$^{\ddag}$ & 21.17 & 0.648 & 48.56 & 10.93 \\
      UDIS2~\cite{nie2023parallax}$^{\ddag}$ & 25.43 & 0.838 & 48.09 & \underline{6.11} \\
      DHS~\cite{mei2024dunhuangstitch}$^{\ddag}$ & \underline{25.88} & \underline{0.845} & \underline{45.73} & 6.18 \\
      \midrule
      \textbf{Ours} & \textbf{25.97} & \textbf{0.852} & \textbf{42.12} & \textbf{5.78} \\
      \bottomrule
    \end{tabular}
    \caption{\textbf{Quantitative comparison on the UDIS-D~\cite{nie2021unsupervised} Dataset.} PSNR/SSIM measure alignment accuracy on overlapping regions, while PIQE/NIQE assess the perceptual quality of the panorama.}
    \label{tab:psnr_ssim_piqe_niqe}
  \end{minipage}

\end{table*}

\subsection{Training}
\label{sec:Training}

\textbf{Training Data.} We collected a large-scale, labeled image stitching dataset named \textbf{Pano360}. This dataset consists of 200 unique real-world scenes, categorized into three types: general tourism (50\%), extreme sports (e.g., skiing and skydiving, 30\%), and challenging illumination conditions (e.g., nightscapes and strong backlighting, 20\%).
For each scene, we captured image sequences at three distinct focal lengths. Each image is annotated with its ground-truth camera parameters. We augment the dataset by applying a random rotational jitter of up to 2 degrees along the yaw, pitch, and roll axes.
Each scene comprises 72 images (24 frames for each of the three focal lengths), spanning a complete 360-degree field of view. All images have a resolution of 2048 × 2048, bringing the total dataset to the size of 14,400 frames.

\begin{figure}[t]
    \centering

    \includegraphics[width=1.0\linewidth]{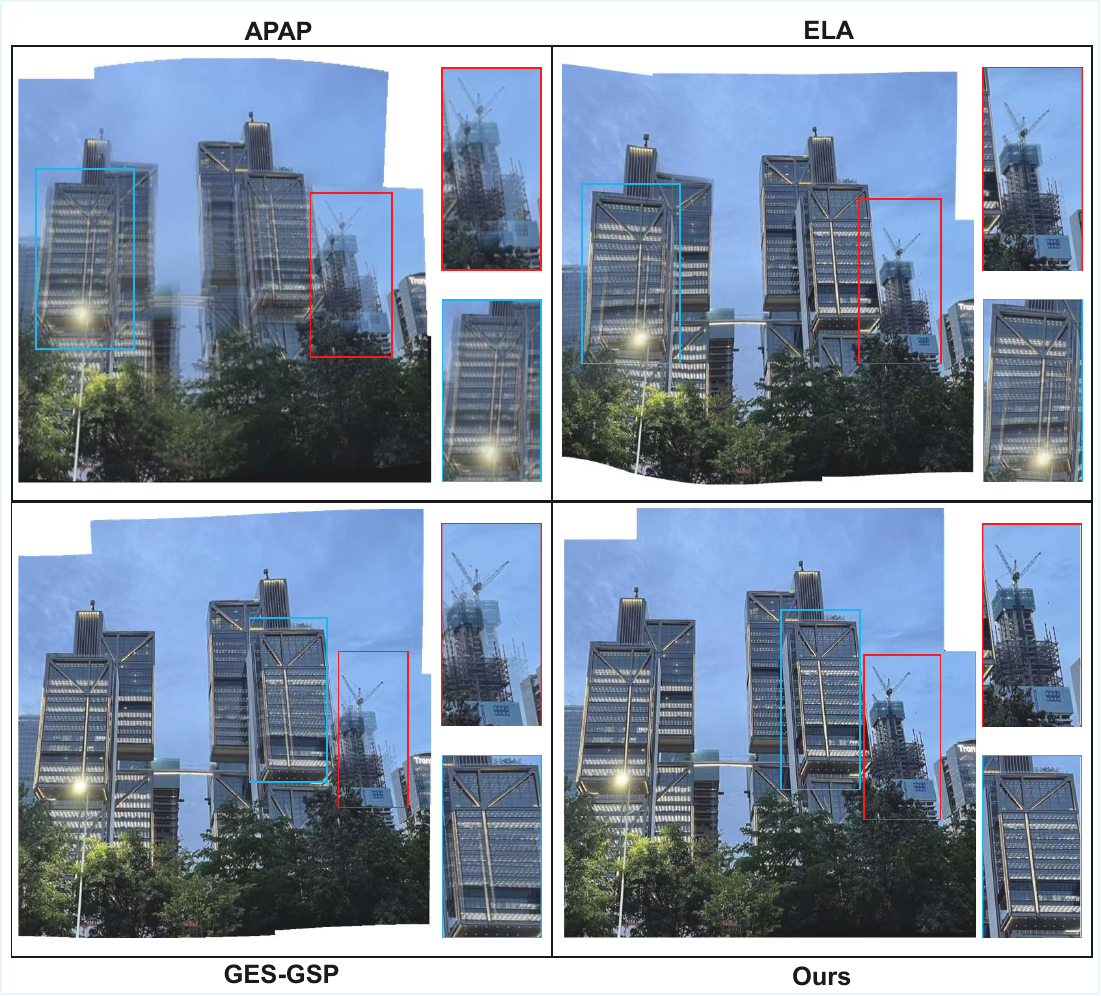} 
    
    \caption{\textbf{Qualitative comparison of panorama stitching on challenging in-the-wild images.} Existing methods suffer from noticeable artifacts and distortions on buildings with repetitive patterns. However, our method achieves globally consistent alignment and remains free of these artifacts.}
    \label{fig:compare}
\end{figure}

\noindent\textbf{Model Training}. We train the Pano360 model by minimizing a multi-task loss. The camera loss $\mathcal{L}_{\text{cam}} $ supervises the camera parameters $\hat{g_i}$ by minimizing the Huber loss between the predicted $\hat{g_i}$ and the ground-truth camera parameters $g_i$:$\mathcal{L}_{\text{cam}} = \sum_{i=1}^{N} \|\hat{\mathbf{g}}_i - \mathbf{g}_i\|_{\epsilon}$. The seam loss $\mathcal{L}_{\text{seam}}$ is computed as the L1 distance between the predicted mask $\hat{M_i}$ and the pseudo-label $M_i$:$\mathcal{L}_{\text{seam}} = \sum_{i=1}^{N} \|\hat{M}_i - M_i\|$. To ensure the continuity of the seamline between images, we exclude the uncertainty terms from the loss function. It allows our model to converge more rapidly. Crucially, the $\mathcal{L}_{\text{proj}} $ enables the network to adapt to different projection formats using predicted camera parameters. It is predefined from the beginning of training to ensure gradient continuity. Additionally, we initialize the weights of the alternating attention module from a pre-trained VGGT model, keeping the module frozen throughout the training process.

\noindent\textbf{Ground Truth Normalization}. The final panorama should be invariant to the permutation of input images, meaning that different orders should produce the same result. Therefore, we solve this problem by normalizing the data, ensuring the transformer network produces a consistent result. We express all quantities in the coordinate system of the first frame and generate the seam mask labels for all images using the method described in \cref{sec:Seam head}. More importantly, we make no assumptions about the output; instead, we force the transformer network to learn the normalization method we chose for the training data.

\section{Experiments}
In this section, we compare our method to state-of-the-art approaches in various scenes to show its generalization and effectiveness.

\subsection{Quantitative Comparison}

We evaluate our method's overall performance on the Pano360 dataset. For each scene, we randomly sample 10 images and recover their panoramas. Following~\cite{feng2025dit360}, we report perceptual quality using traditional no-reference metrics, BRISQUE~\cite{mittal2012no} and NIQE~\cite{mittal2012making}, as well as the modern Q-Align~\cite{wu2023q} score. Q-Align utilizes a multimodal large language model to assess the quality and aesthetics of the final panoramic images.

As presented in \cref{tab:Pano-360_comparison}, our model achieves state-of-the-art across all metrics. It demonstrates that our model has excellent alignment, low-distortion, weak-artifact, and smooth seams. Our method significantly outperforms competing approaches in both success rate and runtime, as shown in \cref{tab:geometric_feature_comparison}. Compared to concurrent works~\cite{brown2007automatic,zaragoza2013projective,li2017parallax,du2022geometric}, which rely on handcrafted geometric features that may fail in areas with weak texture and repetitive patterns, our model demonstrates significant advantages for feature extraction. In contrast to methods~\cite{jia2021leveraging,nie2023parallax} (indicated by $^{\ddag}$) , our model avoids the expensive post-optimization steps for multi-image stitching. By leveraging a novel network to guide images warping in 3D space, our approach shows significantly potential for large-scale stitching scenes.

To validate the generalization capability of our model, we evaluate its performance on the UDIS-D~\cite{nie2021unsupervised} dataset, as shown in \cref{tab:psnr_ssim_piqe_niqe}. Although our model was not trained on this dataset and specifically designed for pairwise stitching, its alignment accuracy is comparable to other methods, which were particularly fine-tuned on it. The perceptual quality of our method is significantly superior to all other alternatives.

\begin{figure*}[t]
    \centering

    \includegraphics[width=1.0\linewidth]{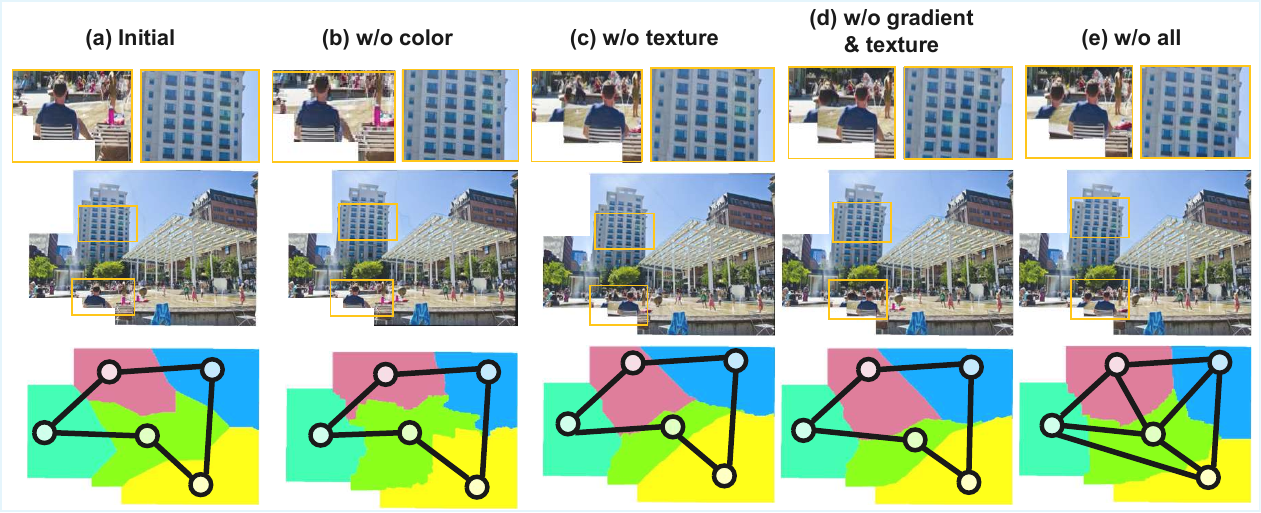} 
    
    \caption{\textbf{Ablation study for seam detection.} (a) Our full method. (b-d) Ablation of different components. (e) Baseline using traditional graph-cut with pairwise computation. Methods (a-d) perform joint optimization strategy for the seam mask.}
    
    \label{fig:seamable} 
\end{figure*}

\subsection{Ablation Studies}
We conduct ablation studies to assess each component’s contribution, using homography-based perspective warping and graph-cut seam detection as the baseline. First, we use LightGlue~\cite{lindenberger2023lightglue} for keypoint detection and matching. Then, the homography transformation is estimated via an 8-DOF algorithm. We evaluate three key modules: pose-guided image warping, projection function, and seam mask detection. All experiments are evaluated on Scene (c) of the Pano360 dataset, as shown in \cref{tab:Ablation_componet}. We report the overall perceptual quality of the panorama as evaluation metrics.

Our process involves three steps. First, we perform an initial alignment by warping each image to the reference view using a homography, which is computed from the corresponding camera parameters: $\mathbf{H}_{i \rightarrow j} = \mathbf{K}_j \mathbf{R}_j \mathbf{R}_i^{T} \mathbf{K}_i^{-1}$. Second, we apply the non-perspective projection format detailed in \cref{sec:Projection Head}. Finally, our proposed approach replaces the traditional graph-cut algorithm for seamless blending.

\begin{table}[t]
  \centering
  \footnotesize
  \setlength{\tabcolsep}{6pt} 
  \renewcommand{\arraystretch}{1.0} 
\begin{tabular}{cccccc}
\toprule
w. $\mathcal{L}_{\text{cam}}$ & w. $\mathcal{L}_{\text{proj}}$ & w. $\mathcal{L}_{\text{seam}}$ & $\text{QA}_{\text{quality}}$ $\uparrow$ & BRIS $\downarrow$ & NIQE $\downarrow$ \\
\midrule
\ding{55} & \ding{55} & \ding{55} & 2.76 & 62.47 & 5.31 \\
\checkmark & \ding{55} & \ding{55} & 3.45 & 47.43 & 4.65 \\
\checkmark & \checkmark & \ding{55} & 3.68 & 43.71 & 3.97 \\
\ding{55} & \ding{55} & \checkmark & 3.01 & 51.12 & 4.83 \\
\checkmark & \checkmark & \checkmark & \textbf{4.09} & \textbf{37.96} & \textbf{3.37} \\
\bottomrule
\end{tabular}
\caption{\textbf{Ablation study for panorama stitching components}, which shows that combining with pose-guided image warping, projection function, and seam mask detection yields the highest performance on Pano360.}
\label{tab:Ablation_componet}
\end{table}

As shown in \cref{tab:Ablation_componet}, pose-guided image warping overcomes the limitations of pairwise homography estimation. Given a batch of input images, it directly predicts the camera parameters for each image, thereby avoiding the accumulated errors in pairwise processing and significantly improving perceptual quality. The projection function mitigates distortions in the panorama via non-perspective mapping. Seam mask detection combines multiple energy terms to generate seamless and perceptually pleasing panoramas. Precise alignment is critical for seam detection, while misalignment leads to visual artifacts in the overlapping region.

To further demonstrate the effectiveness of our seam detection method, we conduct a qualitative comparison in a dynamic outdoor scene, as shown in \cref{fig:seamable}. First, we ablate the color difference term, which leads to noticeable color discrepancies along the seam. Next, we ablate the texture map that guides the gradient computation, causing the seam to cut through a person and produce ghosting artifacts. When we ablate both terms, these artifacts are exacerbated, and significant structural distortion appears on the building. Finally, the traditional graph-cut method produces even more severe distortions. In contrast, our full model, which integrates all components, generates the optimal seam and is free of visual artifacts.

\section{Limitations}
Although our method provides a novel panorama stitching framework and exhibits strong generalization and exceptional performance across diverse in-the-wild scenes, several limitations remain. First, the current model does not support input images with inherent distortions. Second, for scenes with extremely large parallax, such as when the same object is captured from significantly different angles, stitching is not achievable without  3D reconstruction.

\section{Conclusions}
In this paper, we revisit the limitations of both traditional geometric feature-based and learning-based approaches in panorama stitching. We propose a unified framework that preserves  the globally geometric consistency in 3D space. Our method Pano360 demonstrates superior performance in terms of alignment accuracy and perceptual quality across extensive experiments on diverse indoor, outdoor, and aerial scenes. Our simple, efficient and neural network-based approach significantly differs from prior alternatives, which overlook the underlying 3D projective geometry and need expensive post-processing. Our work provides new insights to panoramic vision technologies.

\noindent \textbf{Acknowledgment.} This work is supported by the National Natural Science Foundation of China (32572189), National Key Research and Development Program of China (2022YFF0607001), Shaoguan Science and Technology Research Project (230316116276286), Foshan Science and Technology Research Project (2220001018608), Zhuhai Science and Technology Research Project (2320004002668), Zhongshan Science and Technology Research Project (2024A1010), Guangzhou City Science and Technology Research Projects (2023B01J0011), and Guangdong Basic and Applied Basic Research Foundation (2023A1515010993, 24202107190000687).

{
    \small
    \bibliographystyle{ieeenat_fullname}
    \bibliography{main}
}


\end{document}